\documentclass[10pt,twocolumn,letterpaper]{article}

\usepackage{cvpr}
\usepackage{times}
\usepackage{epsfig}
\usepackage{graphicx}
\usepackage{amsmath}
\usepackage{amssymb}
\usepackage{amsmath}
\usepackage{algorithm}
\usepackage[noend]{algpseudocode}
\usepackage{multirow}
\usepackage{pifont}
\usepackage{booktabs}
\usepackage{pdfpages}

\usepackage[pagebackref=true,breaklinks=true,letterpaper=true,colorlinks,bookmarks=false]{hyperref}

\newfloat{algorithm}{t}{lop}

\newcommand{\ltl}{learning-to-learn }
\newcommand{\xmark}{\ding{55}}%
\newcommand{\scarla}{Carla}
\newcommand{\flying}{F3D}
\newcommand{\ssynthia}{Synthia}

\newcommand{\kitti}[1]{KITTI#1}
\newcommand{\synthia}[1]{Synthia#1}
\newcommand{\carla}{Carla}

\def\dispnet{Dispnet}

\newcommand{\SKIP}[1]{}
\newcommand{\calD}{\mathcal{D}}

\newcommand{\calL}{\mathcal{L}}
\newcommand{\calT}{\mathcal{T}}
\newcommand{\lt}{\textbf{L2A}}
\newcommand{\wlta}{\textbf{L2A+WAd}}
\newcommand{\lta}{\textbf{L2A+Ad}}
\newcommand{\sul}{\textbf{SL}}
\newcommand{\sula}{\textbf{SL+Ad}}

\newcommand{\calV}{\mathcal{V}}
\usepackage{enumitem}

\usepackage{authblk}
\makeatletter
\renewcommand\AB@affilsepx{\quad\quad	\protect\Affilfont}
\makeatother

\cvprfinalcopy % *** Uncomment this line for the final submission

\def\cvprPaperID{4522} % *** Enter the CVPR Paper ID here

\ifcvprfinal\fi
\begin{document}

%%%%%%%%% TITLE
%\title{Continuous unsupervised adaptation for deep stereo via meta learning}
\title{Learning to Adapt for Stereo}

%\author{
%	Alessio Tonioni\textsuperscript{1} \hspace{10mm}  Oscar Rahnama\textsuperscript{*2,4} \hspace{10mm}  Tom Joy\textsuperscript{*2}\\
%	Luigi Di Stefano\textsuperscript{1} \hspace{10mm} Thalaiyasingam Ajanthan\textsuperscript{3} \hspace{10mm} Philip Torr\textsuperscript{2}\\
%	\\
%	1. CVLAB, University of Bologna, \{name.surname\}@unibo.it \hspace{10mm} 2. TVG, University of Oxford\\
%	3. Australian National University \hspace{10mm} 4. FiveAI 	\hspace{10mm} %* equal contribution\\
%}
%\renewcommand*{\thefootnote}{\fnsymbol{footnote}}
% \footnote{Second two authors contributed equally}
%\footnotetext{\textsuperscript{*}Second two authors contributed equally}
\author[1]{Alessio Tonioni\thanks{Work done while at University of Oxford.}}
\newcommand\ThanksMark{\footnotemark[1]}
\author[2,4]{Oscar Rahnama\thanks{Second two authors contributed equally.}}
\newcommand\CoAuthorMark{\footnotemark[2]}
\author[2]{Thomas Joy\protect\CoAuthorMark}
\author[1]{Luigi Di Stefano}
\author[3]{Thalaiyasingam Ajanthan\protect\ThanksMark}
\author[2]{Philip H. S. Torr}
\affil[1]{University of Bologna}
\affil[2]{University of Oxford}
\affil[3]{Australian National University}
\affil[4]{FiveAI}

\maketitle
%\thispagestyle{empty}

%%%%%%%%% ABSTRACT
\begin{abstract}
Real world applications of stereo depth estimation require models that are robust to dynamic variations in the environment.
Even though deep learning based stereo methods are successful, they often fail to generalize to unseen variations in the environment, making them less suitable for practical applications such as autonomous driving.
In this work, we introduce a ``learning-to-adapt'' framework that enables deep stereo methods to continuously adapt to new target domains in an unsupervised manner.
Specifically, our approach incorporates the adaptation procedure into the learning objective to obtain a base set of parameters that are better suited for unsupervised online adaptation.
To further improve the quality of the adaptation, we learn a confidence measure that effectively masks the errors introduced during the unsupervised adaptation.
We evaluate our method on synthetic and real-world stereo datasets and our experiments evidence that learning-to-adapt is, indeed beneficial for online adaptation on vastly different domains.
\end{abstract}

%%%%%%%%% BODY TEXT
\section{Introduction}
\label{sec:intro}
Stereo correspondence estimation is one of the standard methods for predicting the depth of a scene. 
%Given a pair of calibrated images, the objective is to estimate a dense disparity map between them.
State-of-the-art algorithms treat stereo as a supervised learning problem and employ deep convolutional neural networks (CNNs) to directly predict the disparity values~\cite{mayer2016large}.
However, the inability of deep stereo methods to generalize to new domains~\cite{pang2018zoom,Tonioni_2017_ICCV} presents a serious problem in applications where stereo vision is most useful.
Consider an autonomous car driving along the twisting turns, endless meanders and through the frequent tunnels around Lake Como. 
With few or ineffective barriers offering safety from the shear cliffs, it is imperative that the autonomous car performs flawlessly.
Moreover, when passing through frequent tunnels where the lighting conditions change dramatically, a learned stereo system might fail to perform in the expected manner, potentially leading to fatal consequences. 

Fine tuning a learned model on the target environment may help to achieve good performance. However, acquiring real dense ground truth data for stereo is extremely challenging, even with expensive equipment and human effort~\cite{KITTI_2015}. 
Moreover, considering the above example, one cannot expect to collect ground truth data for all possible seasons, times of the day, weather conditions, \etc.
To address this issue, we propose to investigate the use of synthetic data to learn a model offline, which, when deployed, can quickly adapt to any unseen target domain online in an unsupervised manner, eliminating the need for expensive data collection.

We formulate this {\em learning-to-adapt} problem using a meta-learning scheme for continuous adaptation.
Specifically, we rely on a model agnostic meta-learning framework~\cite{maml} due to its theoretical foundation~\cite{finn2017meta}, ease of use, and its successful application on various domains \cite{maml,clavera2018learning,al2017continuous}.
Our goal is to learn a model offline using synthetic data, which can continuously adapt to unseen real video sequences in an unsupervised manner at test time. 
This means our model is always in training mode and its parameters are automatically tuned to the current environment online without the need for supervision. 
Such an online adaptation scenario has been considered previously in the literature~\cite{Tonioni2018real,zhong2018open}. However, in this work, we explicitly \emph{learn-to-adapt} which allows us to achieve superior performance.

\begin{figure*}
	\centering
	\setlength\tabcolsep{1pt}
	\begin{tabular}{cccc}
		\includegraphics[width=0.24\textwidth]{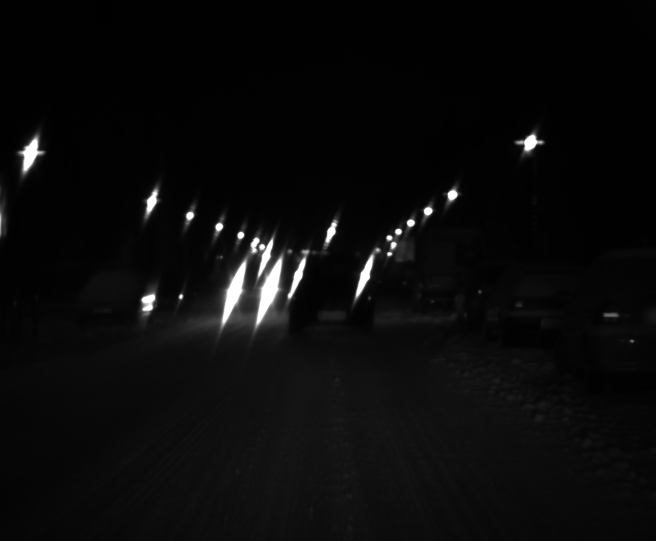} & \includegraphics[width=0.24\textwidth]{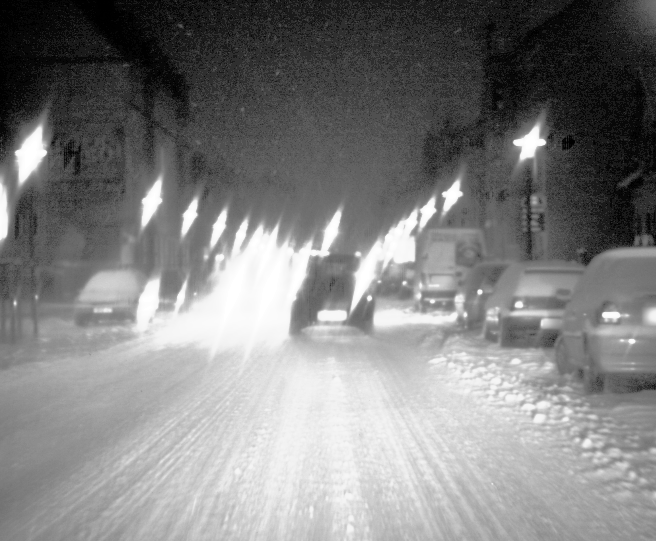} &
		\includegraphics[width=0.24\textwidth]{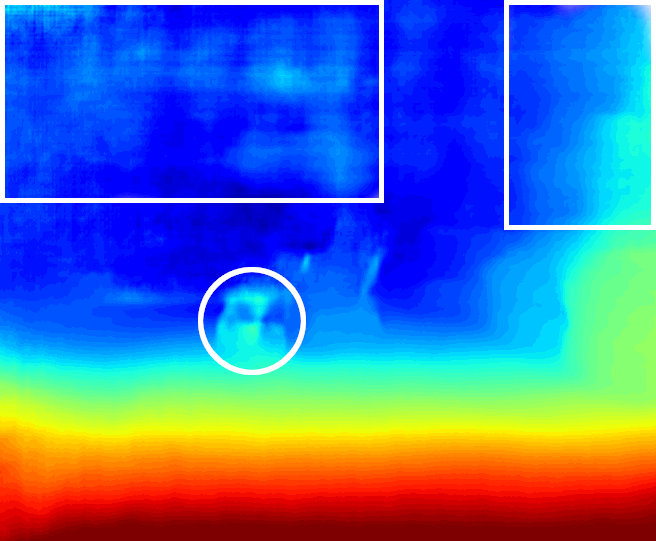} & \includegraphics[width=0.24\textwidth]{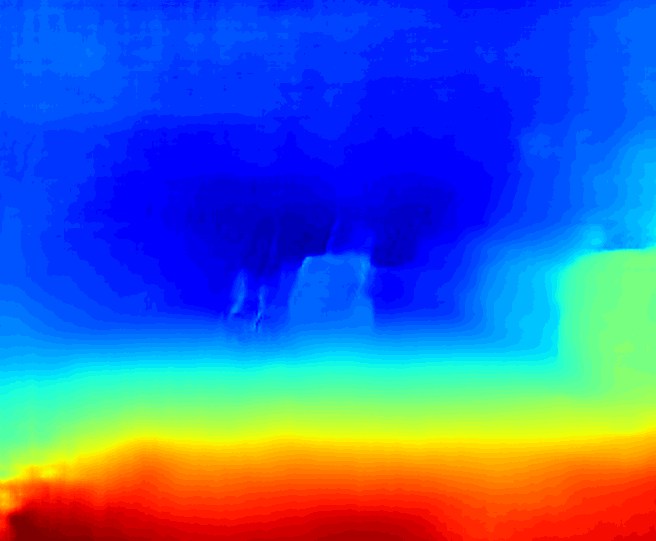}\\
		(a) Left input frame & (b) Left frame equalized & (c) \kitti{} tuned & (d) Ours \\
	\end{tabular}
	\vspace{1ex}
	\caption{We demonstrate the effectiveness of continuous adaptation on a challenging video sequence from \cite{meister2012outdoor}. {\bf(a)} Left input frame, {\bf(b)} histogram equalized frame for visualization purpose, {\bf(c)} disparity map produced by a Dispnet-Corr1D~\cite{mayer2016large} trained on annotated real data from \kitti{}, {\bf(d)} disparity map produced by a Dispnet-Corr1D~\cite{mayer2016large} trained on synthetic data using our learning-to-adapt framework and continuously adapted on this video sequence. The prediction of our method does not suffer from the same artifacts as {\bf (c)} (highlighted in white), thus illustrating the advantage of continuous unsupervised adaptation. }
	\label{fig:teaser}
\end{figure*}

Our meta-learning approach directly incorporates the online adaptation step into the learning objective, thus allowing us to obtain a base set of weights that are better suited for unsupervised online adaptation.
However, since the adaptation is performed in an unsupervised manner (\eg, based on re-projection loss~\cite{garg2016unsupervised,godard2017unsupervised}), it is inherently noisy, causing an adverse effect on the overall algorithm.
To alleviate this deficiency, we learn a confidence measure on the unsupervised loss and use the confidence weighted loss to update the network parameters.
This effectively masks the noise in the adaptation step, preventing detrimental parameter updates.
In our case, the confidence measures are predicted using a small CNN which is incorporated into the meta-learning framework, allowing the network to be trained end-to-end with no additional supervision.

%Furthermore, the overall system is trained end-to-end with no additional supervision. 

In our experiments, we make use of a synthetic stereo dataset (\synthia{}~\cite{Ros_2016_CVPR}), a real-world dataset (\kitti{-raw}~\cite{Uhrig2017THREEDV}),
and generate a new synthetic dataset containing multiple sequences of varying weather and lighting conditions using the Carla simulator~\cite{carla}. 
We evaluate our algorithm between two pairs of dataset domains: 1) Carla to \synthia{}; and 2) Carla or \synthia{} to \kitti{-raw}. 
%1) training on a synthetic dataset and testing on a different synthetic dataset; and 2) similarly training on a synthetic dataset and testing on a real dataset. 
In all experiments, our learning-to-adapt method consistently outperforms previous unsupervised adaptation methods~\cite{Tonioni2018real}, validating our hypothesis that learning-to-adapt provides an effective framework for stereo.

%%%%%%%%%%%%%%%%%%%%%%%%%%%%%%%%%%%%%%%%%%%%%%%%%%%%%%
\section{Problem Setup and Preliminaries}
\label{sec:preliminaries}
In this section, first we formalize online adaptation and discuss its advantages. We then briefly review a meta-learning algorithm which we will transpose into our continuous adaptation scenario.

\subsection{Online Adaptation for Stereo}
\label{sec:adaptation}
Let us denote two datasets of stereo video sequences: $\calD_s$ (supervised), with available ground truth, and $\calD_u$ (unsupervised), without ground truth.
We would like to learn network parameters offline using $\calD_s$, and use $\calD_u$ as the target (or test) domain.
However, in contrast to the standard evaluation setting and following the evaluation protocol of \cite{Tonioni2018real,zhong2018open}, the model is allowed to adapt to the target domain in an unsupervised manner.
We follow the online adaptation paradigm proposed in \cite{Tonioni2018real}, \ie, for each new frame acquired we perform a single gradient descent step to keep the optimization fast and allow better handling of a rapidly changing test environment.

Formally, let the parameters of the base model trained on $\calD_s$ be $\theta$.
Given an unseen target video sequence ${\calV\in \calD_u}$, the adaptation is iteratively performed for each consecutive stereo pair in the sequence, using gradient descent on a predefined unsupervised loss function ($\calL_u$).
At iteration $t$, the online adaptation can be written as:
 \begin{equation}\label{eq:update}
{\theta_t} \gets {\theta}_{t-1} - \alpha \nabla_\theta \mathcal{L}_u(\theta_{t-1},i_t)\ ,\\
\end{equation}
where $\theta_0 = \theta$, $\alpha>0$ is the learning rate and $i_t$ denotes the stereo pair of $t^{th}$ frame of the sequence $\calV$.
Note that the network parameters are {\em continuously} updated for the entire video in a sequential manner.
This process is repeated for each video sequence starting from the base model  $\theta$.

\paragraph{Motivating Example.}
%Online adaptation for stereo is realistic as it is impossible to anticipate all the possible target domains and collect data beforehand.
To show that deep CNN based stereo networks are highly sensitive to domain-shift and that online adaptation is indeed necessary, we give a motivating example as follows.
We select a video sequence from \cite{meister2012outdoor} as a test domain, where the environment is similar to that of \kitti{} but features extreme weather conditions (\eg, night, snow, \etc).
We compare the predicted disparities of a Dispnet-Corr1D network~\cite{mayer2016large} for two training regimes.
The first is fine-tuned on the \kitti{} training sets~\cite{KITTI_2012,KITTI_2015}, and the second is trained on synthetic data using our learning-to-adapt framework and performs unsupervised online adaptation for the target domain.
The results are shown in \autoref{fig:teaser}. Here it is evident that fine tuning on real images is not sufficient to obtain reliable performance across all environments as evidenced by the major mistakes in (c) marked in white. 
As can be seen, (c) behaves worse than the network trained only on synthetic data and adapted online to the target domain in an unsupervised manner by our formulation (d).

\subsection{Model Agnostic Meta Learning}
\label{sec:maml}
Model Agnostic Meta Learning (MAML)~\cite{maml} is a popular meta-learning algorithm designed for few-shot learning problems.
%The main idea is to include the test time setup into the training paradigm.
The objective is to learn a base model $\theta^*$, which enables fast adaptation to new tasks when used as initial weights.
This is achieved by forming a nested optimisation problem, where, in the inner loop, we perform SGD for each task in the standard way.
In the outer loop, the base model parameters are optimized using the loss of all the tasks, enabling fast adaptation.
%we optimise the initial starting weights for fast adaptation.

Let $\calT$ be the set of tasks in the training set and let the task-specific training and validation sets be $\mathcal{D}_{\tau}^{\mbox{\emph{train}}}$ and $\mathcal{D}_{\tau}^{\mbox{\emph{val}}}$ respectively for $\tau \in \calT$.
Assuming a single gradient step in the inner loop, 
the overall MAML objective can be written as: 
\begin{equation}
    \label{eq:maml}
    \min_{\theta} \sum_{\tau \in \mathcal{T}} \mathcal{L}(\theta - \alpha \nabla_\theta \mathcal{L}(\theta,\mathcal{D}_{\tau}^{\mbox{\emph{train}}}),\mathcal{D}_{\tau}^{\mbox{\emph{val}}}), 
\end{equation}
where $\alpha>0$ is the learning rate used for adaptation.
As previously stated, this meta-objective function is optimized via a two-stage gradient descent algorithm.
Specifically, at each optimization iteration, the inner-loop performs a gradient descent update for each task separately starting from a common base model $\theta$ (adaptation step).
Then, the outer-loop performs an update on the common base model, where the gradient is the sum of task-specific gradients computed using the parameters updated in the inner loop.
We refer the interested reader to the original paper for more detail~\cite{maml}.

\section{Learning to Adapt for Stereo}
\label{sec:method}

We first design a meta-learning algorithm for stereo adaptation by incorporating unsupervised continuous adaptation into the training paradigm.
Then, we introduce a new mechanism to re-weight the pixel-wise errors estimated by the unsupervised loss function, making the adaptation more effective.

\begin{algorithm}[t]
\caption{Adaptation at training time for sequence $\calV^\tau$}
\label{alg:adapt}
    \textbf{Require:} $\theta, \calV^{\tau} = [i^\tau_1,\dots,i^\tau_n]$
    \begin{algorithmic}[1]
        \State $\theta^{\tau}_0\gets \theta$\Comment{Parameter initialization}
                \For{$t \gets 1,\ldots,n-1$}
                    \State $\theta^{\tau}_t \gets \theta^{\tau}_{t-1} - \alpha \nabla_{\theta^{\tau}_{t-1}} \mathcal{L}_u\left(\theta^{\tau}_{t-1},i_t\right)$\Comment{Adaptation}
                    \State $\calL_s\left(\theta^{\tau}_{t}, i^{\tau}_{t+1}\right)$\Comment{Supervised evaluation}
                \EndFor
    \end{algorithmic}
\end{algorithm}

\subsection{Meta Learning for Stereo Adaptation}
\label{ssec:method_meta}

\begin{figure*}[t]
    \centering
    \includegraphics[width=1\textwidth]{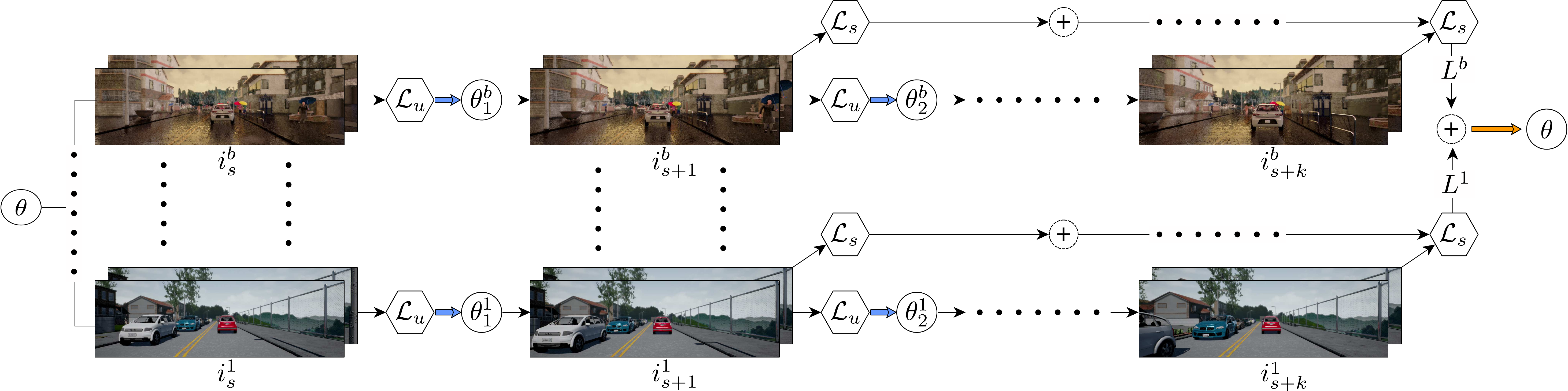}
    \caption{Representation of one iteration of meta-learning of network parameters $\theta$ using a batch of $b$ sequences and sampling $k$ frames from each. 
    We represent loss computation steps with hexagons and gradient descent steps with colored arrows. 
    Blue and orange arrows denote adaptation steps and meta-learning steps, respectively. 
    Starting from an initial set of parameters $\theta$, the network is adapted independently on each sequence using loss function $\calL_u$. 
    The adapted models are evaluated on the following frame of each sequence using loss function $\calL_s$. 
    Finally, the initial parameters $\theta$ are updated via a gradient descent to minimize the sum of loss functions obtained by all evaluated models.}
    \label{fig:meta_learn} 
\end{figure*}

Our hypothesis is that for any deep stereo network, before performing online adaptation to a target domain, it is beneficial to learn a base set of parameters ($\theta$) that can be adapted quickly and effectively to unseen environments. 
We observe that our objective of learning-to-adapt to unseen video sequences is similar in spirit to that of MAML. %which tackled different tasks.
Here, we perform the single task of dense disparity regression, but learn how to adapt to different environments and conditions.

% being challenged by new tasks we are challenged by different conditions and environments . %across many video sequences. %To that extent, we train a set of parameters for a deep stereo network to learn how to efficiently adapt to new sequences.

We model an environment through a stereo video sequence $\calV^\tau=[i^\tau_1,\dots,i^\tau_n]$\footnote{For simplicity we assumed that all video sequences are of same length, but this is not a necessity.}.
At test time, the parameters are continuously adapted to the sequence  $\calV^\tau$ in an unsupervised manner according to \autoref{eq:update}.
At training time, we mimic the same adaptation process on training sequences and evaluate the performance of the model after each adaptation step on the subsequent frame. 
To measure the performance, we rely on a supervised loss function $\calL_s$ (\eg, $L_1$ or $L_2$ regression). This procedure for a single sequence $\calV^\tau$ is given in \autoref{alg:adapt}.  

During training we perform this adaptation on a supervised training set of video sequences $\calD_s$ (\eg, a set of rendered synthetic video sequences). The final objective of our problem is to maximise the measured performance across all frames and all sequences in $\calD_s$. This can be written in a compact form as:

\begin{equation}
    \label{eq:loss_meta}
    \min_{\theta} \sum_{\calV^\tau \in \calD_s} \sum_{t=1}^{n-1} \mathcal{L}_s(\theta^\tau_t,i^\tau_{t+1})\ , 
\end{equation}
where $\theta_t^\tau$ is obtained sequentially through updates as detailed in \autoref{alg:adapt}. 

Note that this formula extends \autoref{eq:maml} (MAML) to the continuous and unsupervised adaptation scenario. Contrary to \autoref{eq:maml}, we use two different loss functions: 1) an unsupervised loss ($\mathcal{L}_u$) to adapt a model to a video sequence; and 2) a supervised loss ($\mathcal{L}_s$) for the optimization of the set of parameters $\theta$. 
We make this distinction to mimic the test time behaviour. 
Specifically, $\mathcal{L}_u$ (\ie, some form of unsupervised loss function) will be used at test time, while $\mathcal{L}_s$ can use all the available annotations of the training set for optimization. 
Our intuition is that by using two different loss functions, $\theta$ can be optimized such that it is better suited to be adapted without supervision (\ie, by $\mathcal{L}_u$), while the performance is measured with respect to the ground truth (\ie, by $\mathcal{L}_s$). 

\begin{algorithm}[t]
\caption{Learning to Adapt for Stereo}
\label{alg:meta}
    \textbf{Require:} Training set $\calD_s$, and hyper-parameters $\alpha,\beta,k,b$
    
    %\textbf{Require:} $\alpha,\beta,k$  hyper-parameters.
    \begin{algorithmic}[1]
        \State Initialize $\theta$
        \While{$not$ $done$}
            \State $\calD^b\sim \calD_s$ \Comment{Sample a batch of sequences}
            \ForAll{$\calV^{\tau}\in\calD^b$}
                \State $\theta^{\tau} \gets \theta$ \Comment{Initialize model} 
                \State $L^{\tau} \gets 0$ \Comment{Initialize accumulator}
                \State $[i_s,\dots,i_{s+k}] \sim \calV^{\tau}$ \Comment{Sample $k$ frames}
                \For{$t \gets s,\ldots,s+k-1$}
                    \State $\theta^{\tau} \gets \theta^{\tau} - \alpha \nabla_{\theta^{\tau}} \mathcal{L}_u(\theta^{\tau},i_t)$ \Comment{Adaptation}
                    \State $L^{\tau} \gets L^{\tau} + \mathcal{L}_s(\theta^{\tau},i_{t+1})$ \Comment{Evaluation}
                \EndFor
            \EndFor
            \State $\theta \gets \theta - \beta \nabla_\theta \sum_{\calV^{\tau}\in\calD^b} L^{\tau}$ \Comment{Optimization}
        \EndWhile
    \end{algorithmic}
\end{algorithm}

Note that optimizing \autoref{eq:loss_meta} on complete video sequences would be infeasible for long video sequences as the memory requirement grows linearly with $n$. 
To alleviate this, we approximate it by optimizing over batches of sequences of $k$ randomly sampled consecutive frames. 
Our meta-learning algorithm is detailed in \autoref{alg:meta}.
After sampling a batch of sequences (line 3) and $k$ random frames from each sequence (line 7), we perform unsupervised adaptation on the current frame (line 9) and measure the effectiveness of this update on the following frame (line 10).
This process is repeated for $k$ frames. 
Finally, we optimize the base model parameters $\theta$ to minimize the sum of the supervised losses computed across all the sequences and all the frames (line 11). 
Here, $\alpha$ and $\beta$ are the two learning rates used for online adaptation and for meta training, respectively. 
In \autoref{fig:meta_learn}, we illustrate one optimization iteration of the network parameters $\theta$ using a batch of $b$ sequences and $k$ frames from each.

By optimizing \autoref{eq:loss_meta} we are able to learn a base parameter configuration $\theta$ suited for adaptation. 
However, the use of an imperfect unsupervised loss function ($\mathcal{L}_u$) for adaptation introduces mistakes in the optimization process that may have an adverse effect on the overall algorithm.
To alleviate this issue, we introduce a mechanism to learn to recognize the noise (or mistakes) in the unsupervised loss estimation which can then be masked effectively.

\subsection{Confidence Weighted Adaptation}
\label{ssec:method_reweighting}

\begin{figure}
    \centering
    \includegraphics[width=\columnwidth]{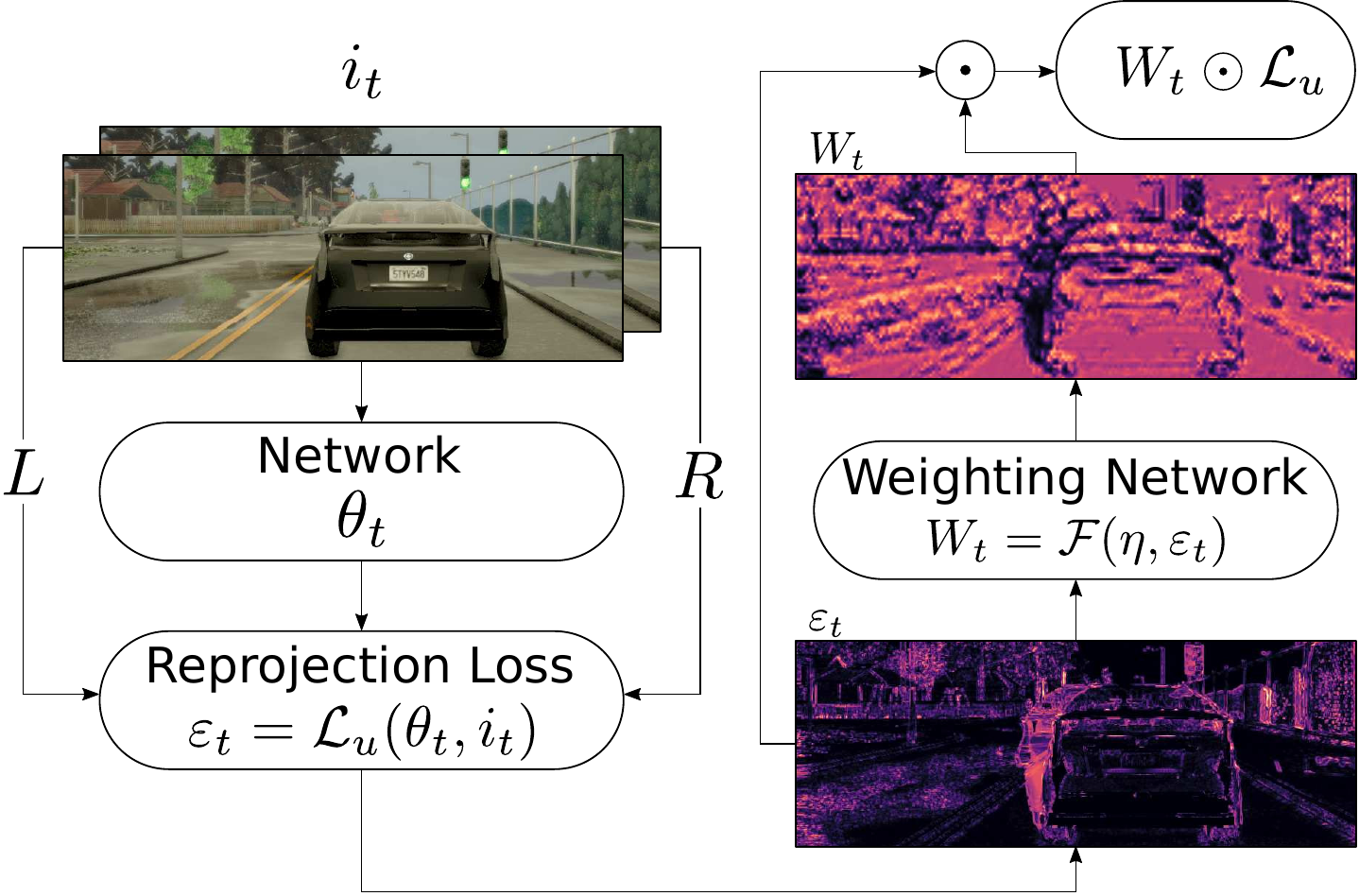}
    \caption{Schematic representation of our weighted adaptation for a single stereo frame using an unsupervised re-projection based loss function $\calL_u$ (bright colors indicate higher values). 
    The system takes a stereo-pair ($i_t$) and computes a disparity map as well as a re-projection loss ($\varepsilon_t$). 
    This loss is then weighted according to $W_t$ effectively masking the mistakes.}
    \label{fig:weighting}
\end{figure}

Unsupervised loss functions for dense disparity estimation often compute some form of pixel-wise error map and minimize over the average mistake. 
Unfortunately, this process is not perfect and usually introduces errors in the optimization process. This may result in sub-optimal performance when compared to the use of supervised loss functions. 
For example, the left-right re-projection loss proposed in~\cite{garg2016unsupervised} is well-known to produce mistakes in occluded areas and reflective surfaces. 
These mistakes are not due to bad predictions by the disparity estimation model, but instead are due to differences between the left and right frames.
Ideally, we would like to have a confidence function to detect erroneous estimations of this loss such that they can be effectively masked. %and  a cleaner loss optimize using a cleaner loss estimation. 
However, training such a confidence function might be difficult since there is no easy procedure to obtain ground-truth annotations for this task. 
We propose to avoid explicit supervised training, and instead, automatically learn to detect noise in the loss estimations by incorporating this new objective into our meta-learning formulation.

In particular, we propose to learn a small CNN that takes a pixel-wise error map estimated by $\mathcal{L}_u$ as an input and produces a tensor $W$ as an output, which has the same shape as the input and its elements are between $0$ and $1$. 
This output can be interpreted as a per pixel confidence on the reliability of the loss estimation with $1$ corresponding to high reliability. 
We can now mask out potentially erroneous estimations by multiplying the loss values by their corresponding confidence values. 
The result is a cleaner measurement of the loss function that reduces detrimental weight updates due to incorrect loss values. 
The idea of masking or weighting the contribution of single examples in the presence of noise or class imbalance in the labels has been previously studied for supervised classification in~\cite{jiang2017mentornet,ren2018learning}. 
In our case, we transpose the similar idea to pixel-wise loss, estimated for a dense regression task and directly predict a dense confidence map. 

Let $W=\mathcal{F}(\eta,\varepsilon)$ be the mask produced by the re-weighting network parametrized by $\eta$ and $\varepsilon=\mathcal{L}_u(\theta,i)$ be the estimated pixel-wise error map computed on the prediction of the disparity model with parameter $\theta$ on stereo frame $i$. 
We normalize the elements of $W$ by dividing each one of them by the number of elements in $W$. 
Now, by modifying \autoref{eq:update}, the final weighted adaptation formula can be written as:
\begin{equation}
    \label{eq:weighted_update}
    \begin{split}
    \widetilde{\theta_t} & \gets \widetilde{\theta}_{t-1} - \alpha \nabla_{\widetilde{\theta}}  \left(W_{t} \odot \mathcal{L}_u(\widetilde{\theta}_{t-1},i_t)\right)\ ,\\
    W_t & = \mathcal{F}(\eta,\mathcal{L}_u(\widetilde{\theta}_{t-1},i_t))\ .
    \end{split}
\end{equation}
where $\widetilde{\theta}_0=\theta$ and $\odot$ indicating the element-wise product between the matrices. 
Note that, the dimension of $\widetilde{\theta}$ is the same as that of $\theta$, however we denote it differently to highlight the fact that $\widetilde{\theta}$ depends on both the base model ($\theta$) as well as the re-weighting network ($\eta$).
%We denote $\widetilde{\theta}$ as the network parameters which are updated according to the unsupervised loss weighted by $W$.

In \autoref{fig:weighting}, we show a schematic representation of our proposed weighted adaptation computed from a single stereo input frame $i_t$. 
On the bottom right corner we give a visualization of the error map produced by an unsupervised re-projection loss $\mathcal{L}_u$, while the top right corner shows a possible confidence mask $W_t$. 
In this example the weighting network is masking errors due to occlusions (\eg, on the left side of the car) and due to reflections (\eg, the puddles on the road).   

Since supervision is not available for $W$, we indirectly train $\eta$ by incorporating \autoref{eq:weighted_update} inside the meta-learning objective described in \autoref{eq:loss_meta}. The final objective of our complete system becomes:
\begin{equation}
    \label{eq:loss}
    \min_{\theta,\eta} \sum_{\calV_\tau \in \mathcal{D}_s} \sum_{t=1}^{n-1} \mathcal{L}_s(\widetilde{\theta}^\tau_t,i^\tau_t)\ .
\end{equation}
Here, $\widetilde{\theta}^\tau_t$ are the parameters of the model updated according to the weighted unsupervised loss function on sequence $\calV^\tau$. As such it depends on both $\eta$ and $\theta$ according to \autoref{eq:weighted_update}.
The whole network can be trained end-to-end with the only supervision coming from the depth annotations used to compute $\mathcal{L}_s$. 
Both $\theta$ and $\eta$ are tuned to maximize the network performances after few steps of optimization as measured by $\mathcal{L}_s$. 
By optimizing a single objective function we are able to learn the parameters ($\eta$) of the weighting network, and a set of base weights for the disparity estimation model ($\theta$) which allow for fast adaptation.

%%%%%%%%%%%%%%%%%%%%%%%%%%%%%%%%%%%%%%%%%%%%%%%%%%%%%

\section{Related Work}
\label{sec:related}

%We briefly review the recent literature for deep stereo models alongside with methods concerned with unsupervised adaptation for stereo and methods for meta-learning.

\paragraph{Machine Learning for Stereo.} 
Mayer \etal{} \cite{mayer2016large}, proposed the first end-to-end stereo architecture that, despite not having achieved state-of-the-art accuracy, initiated a huge shift in stereo literature towards CNN based models. More recent proposals \cite{Kendall_2017_ICCV,Pang_2017_ICCV_Workshops,liang2018learning,chang2018pyramid,jie2018left} have quickly reached top performance on the challenging \kitti{} benchmarks by deploying 3D convolution \cite{Kendall_2017_ICCV}, two-stage refinement \cite{Pang_2017_ICCV_Workshops} and pyramidal elaboration \cite{chang2018pyramid}. 
All these works share the same training protocol. Specifically, the network is first pretrained on the large and perfectly annotated synthetic FlyingThings3D dataset~\cite{mayer2016large} and then fine tuned on the smaller \kitti{} training sets.  

\vspace{-2ex}
\paragraph{Unsupervised Adaptation for Stereo.} 
Tonioni \etal~\cite{Tonioni_2017_ICCV} highlight how machine learning models for stereo are data dependent and, if exposed to environments different from the ones observed during training, will suffer from a severe loss in performance. 
To overcome this problem they introduce an unsupervised way of adapting networks to new domains by deploying traditional stereo algorithms and confidence measures. 
Pang \etal~\cite{pang2018zoom} achieve the same objective by an iterative optimization of predictions obtained at multiple resolutions, while many recent works \cite{zhou2017unsupervisedStereo,godard2017unsupervised,zhang2018activestereonet,3net18} warp different views according to the predicted disparity and minimize the re-projection error. 

Recently, the adaptation problem has also been addressed through an online learning perspective focusing on inference speed \cite{Tonioni2018real}. 
On a related topic, Zhong \etal~\cite{zhong2018open} propose to use video sequences to train a deep network online from random initialization. Moreover, they employ a LSTM in their model to leverage temporal information during the prediction. Similarly to \cite{Tonioni2018real,zhong2018open}, we constantly train our network when deployed on unseen environments, but we additionally propose to learn a good set of initial weights and a confidence function for the loss function that will improve the adaptation process. 

\vspace{-2ex}
\paragraph{Meta Learning.} 
Meta-learning is a long-standing problem in machine learning \cite{naik1992meta,thrun1998learning,schmidhuber1987evolutionary} that tries to exploit structures in data to learn more effective learning rules or algorithms. 
Most of the recent developments in meta-learning algorithms have focused on the task of few shot classification \cite{vinyals2016matching, snell2017prototypical,Sachin2017}, with few exceptions like \cite{maml, mishra2018simple} extending their models to simple function regression and reinforcement learning. 
In \cite{maml}, the authors propose to constrain the learning rule for the model to be stochastic gradient descent and update the initial weight configuration of a network to make it more suited to learning new tasks. This simple formulation has been recently extended to address online adaptation in reinforcement learning using meta-learning to adapt to changing agents \cite{clavera2018learning} or non-stationary and competitive environments \cite{al2017continuous}. 
Our work builds on \cite{maml} by modifying it to use structured regression, unsupervised loss functions and temporal consistency during the update process.    

%%%%%%%%%%%%%%%%%%%%%%%%%%%%%%%%%%%%%%%%%%%%%%%%%%%%%
\section{Experiments}
\label{sec:experimental}
This section presents an evaluation of the quality of our proposed adaptation method. 
Firstly, we lay out our evaluation setup in \autoref{ssec:experimental_setup}. 
Secondly, in \autoref{ssec:results}, we provide qualitative and quantitative results for two pairs of domains: 
1) synthetic to real (\ie, training on synthetic data and testing on real data from \kitti{}); and 2) synthetic to synthetic (\ie, training on one synthetic dataset and testing on a different synthetic domain). 
Finally, in \autoref{ssec:confidence_qualitative} we report qualitative results illustrating our confidence weighted loss.
We provide the code needed to implement our framework to ease further research in this field \footnote{https://github.com/CVLAB-Unibo/Learning2AdaptForStereo}.

\subsection{Experimental Setup}
\label{ssec:experimental_setup}

%\vspace{-0.5cm}
\paragraph{Datasets.} In our experimental evaluation we simulate realistic test conditions, in which no data from the target domain is available.  
We therefore use training and testing data sampled from two completely disjoint datasets. 
For the real dataset, we use the 71 different sequences of the \kitti{-raw} dataset~\cite{KITTI_RAW} (denoted as \kitti{}) accounting for $\sim$43K images with sparse depth annotations provided by \cite{Uhrig2017THREEDV}.

For the synthetic dataset, we have used the FlyingThings3D dataset \cite{mayer2016large} (shortened \flying{}) to perform an initial training of the networks from random initialization. 
Then, we use Synthia~\cite{Ros_2016_CVPR} as a synthetic dataset containing scenarios similar to \kitti{}. 
The dataset is composed of 50 different video sequences rendered in different seasons and weather conditions for a total of $\sim$45K images. 
For this dataset we scaled the image to half resolution to bring the disparity into the same range as \kitti{}.

Finally, using the \carla{} simulator~\cite{carla}, we have rendered a new dataset (referenced as \scarla{}) composed of 25 different video sequences, each being a thousand frames long, with accurate ground truth data for each frame. 
Each video sequence is rendered in 15 different weather conditions to add variance to the dataset. Resulting in a total of $375$K frames. 
During the rendering we set up the virtual cameras to match the geometry of the real \kitti{} dataset (\ie, same baseline, field of view and similar image resolution).

%\vspace{-0.5cm}
\paragraph{Network Architectures.} 
For the experiments we have selected the Dispnet-Corr1D \cite{mayer2016large} architecture (shortened to \dispnet{}). % and a 
For all the evaluation tests, we pretrain the networks on \flying{} to obtain a set of weights that will be used as an initialization across all the other tests.
%
%To get the full resolution disparity from the low-resolution prediction we upscale by bilinear upsampling. 
We implement the confidence function introduced in \autoref{ssec:method_reweighting} as a small three layer fully convolutional CNN with batch normalization. 
The network takes the re-projection error scaled to quarter resolution as an input and produces an output at the same resolution. 
The prediction is then scaled to full resolution using bilinear upsampling. 
More details on the network architectures and on the hyper-parameters used to pretrain them are reported in the supplementary material. 

%\vspace{-0.5cm}
\paragraph{Evaluation Protocol.} 
After an initial offline training, we perform online adaptation and evaluate models on sequences of stereo frames.
%Given that the premise of this work is to learn-to-adapt, we .
%Thus, providing a protocol where the networks can adapt to the sequence before analysing their performance.
To test independent adaptations for each sequence, we reset the disparity network to its trained weight configuration at the beginning of each test sequence.
%Given that the network is constantly in training mode, we reset the disparity network to its trained weight configuration at the beginning of each test sequence, thus allowing the network to adapt for each sequence.
Then, for each frame, first, we measure the performance of the current model and then we adapt it by a single step of back-propagation and weight update according to \autoref{eq:update} before moving to the next frame.
We do not measure the performance on frames used for adaptation.
 
%\vspace{-0.5cm}
\paragraph{Metrics.} 
We measure performance according to both average End Point Error (EPE) and percentage of pixels with disparity error larger than 3 (D1-all).
 Firstly, we measure both metrics independently for each frame to plot performance as a function of the number of frames processed for adaptation.
 Secondly, we average over each sequence and finally over all the dataset. 

%\vspace{-0.5cm}
\paragraph{Offline Training.}
After the initial pretraining on \flying{} we fine tune the networks on a training set with our learning-to-adapt framework \autoref{alg:meta}, we use $k=3$ consecutive frames for each sample and set the learning rates $\alpha=0.00001$ and $\beta = 0.0001$

%\vspace{-0.5cm}
\paragraph{Online Adaptation.} 
We use the left-right re-projected unsupervised loss~\cite{godard2017unsupervised} for the adaptation. Optimization is performed using gradient descent with momentum, where the momentum value is set to $0.9$ and a learning rate set to $0.0001$.

\subsection{Results}
\label{ssec:results}
We evaluate our \ltl{} method between pairs of datasets, one for training, and one for evaluation. 
We consider two scenarios: 1) synthetic to real and 2) synthetic to synthetic.
We compare the results of our learning-to-adapt framework (\lt), and the method trained using a supervised $L_1$ regression loss (\sul). 
Methods performing unsupervised online adaptation at test time are indicated by appending \textbf{+Ad} to the training method, and confidence weighted adaptation by \textbf{+WAd}.
It is worth noting that \sula{} corresponds to the adaptation method proposed in \cite{Tonioni2018real}.
%To promote the use of continuous adaptation, we also provide results with and without adaptation, indicated by \sula{} and \sul{} respectively. 

\subsubsection{Synthetic to Real}
\label{sssec:kitti}
The most interesting scenario is the one where training on a synthetic domain is followed by testing on a real-life domain. 
Specifically, we train on \ssynthia{} or \scarla{} and then evaluate on the \kitti{} dataset. 

\begin{table}[t]
    \centering
   \scalebox{0.75}{
    \begin{tabular}{l@{\hspace{.8\tabcolsep}}c@{\hspace{.7\tabcolsep}}lcccccc}
        \toprule
        && Method & Training set & D1-all (\%) & EPE & $\Delta$D1 & $\Delta$EPE\\
        \midrule
        &(a) & \sul & - & 9.43 & 1.62 & - & -\\
        &(b) & \sula\cite{Tonioni2018real}  & -  & 7.81 & 1.44 & -1.62 & -0.18 \\
        \midrule
        &(c) & \sul & \scarla{} & 7.46 & 1.48 & - & - \\
        &(d) & \sula\cite{Tonioni2018real}  & \scarla{} & 5.26 & 1.20 & -2.20 & -0.28 \\
        &(e) & \sul & \ssynthia{} &  8.55 & 1.51 & - & - \\
        &(f) & \sula\cite{Tonioni2018real} & \ssynthia{} & 5.33 & 1.19 & -3.22 & -0.32 \\
        \midrule
        \multirow{4}{*}{\rotatebox[]{90}{Ours}}&(g) & \lt & \scarla{} & 8.41 & 1.51 & - & -\\
        &(h) & \wlta & \scarla{} & \textbf{4.49} & \textbf{1.12} & \textbf{-3.92} & \textbf{-0.39}\\
        &(i) & \lt & \ssynthia{} & 8.22 & 1.50 & - & - \\
        &(j) & \wlta & \ssynthia{} & 4.65  & 1.14 & -3.57 & -0.36\\
        \midrule
        &(k) & \sul{} (ideal) & \kitti{} & 4.26 & 1.12 & - &  -\\
        \bottomrule
        
    \end{tabular}
    }
\vspace{3mm}
    \caption{Performance on \kitti{} for the \dispnet{} network trained according to different methods after initialization from \flying{}. 
    It can clearly be seen that online adaptation (\textbf{+Ad}/\textbf{+WAd}) provides a significant improvement compared to when it is left out. 
    The best results are obtained when the training is achieved using our \wlta{} framework. 
    Line (k) indicates an upper bound on how well \dispnet{} can perform when fine tuned on a subset of samples from the target domain. 
    The last two columns indicate the performance improvement with adaptation, and as it is evident in the table, our \wlta{} method obtains the largest increase in performance with adaptation.}
    \label{tab:dispnet_kitti}

\end{table}

The results for the \dispnet{} architecture are provided in \autoref{tab:dispnet_kitti}.
Lines (a) to (f) report the performance when the network weights are obtained in a standard way (using a supervised $L_1$ loss function). 
As expected, the network performs poorly when tested on a different domain with respect to the one it was trained on - lines (a, c, e). 
The use of adaptation for this setup provides a significant improvement - lines (b, d, f) - further motivating the need to adapt to a new domain.

The two rows (h) and (j) report results obtained by learning to adapt on \scarla{} or \ssynthia{} using the \wlta{} framework. 
Our proposed framework clearly outperforms the baseline methods for both training datasets. 
%The best overall performance among the methods - trained with supervision only on synthetic data - is achieved by \wlta{} on \carla{} dataset. 
Comparing lines (h) and (d) clearly shows that our training process is able to learn a model which is better suited for continuous adaptation.
The same conclusions hold even for the results with \ssynthia{} (lines (j) and (f)). 
In the last two columns we can observe the relative improvement provided by adaptation for each method.
In these results, it is evident that our \wlta{} framework provides the largest increase in accuracy when performing adaptation.
Lastly, in line (k), we provide the performance of \dispnet{} obtained in the ideal scenario where samples from the target domains are available (\ie, \kitti{2012} and \kitti{2015} training sets) and used to fine tune the base model with a supervised $L_1$ regression loss.
Although having access to such samples would defeat the purpose of our approach, the result listed here ultimately serves as an upper bound on the attainable performance. 
% Line (k) provides the performance of \dispnet{} obtained by fine tuning the base model using a supervised $L_1$ regression loss on samples from the target domains (\ie, \kitti{2012} and \kitti{2015} training sets).q 
As shown, our \wlta{} framework obtains competitive results.
%, even when compared to fine-tuning on a small dataset from the target domain with accurate ground truths. 

\begin{figure}
	\centering
	\includegraphics[width=0.5\textwidth]{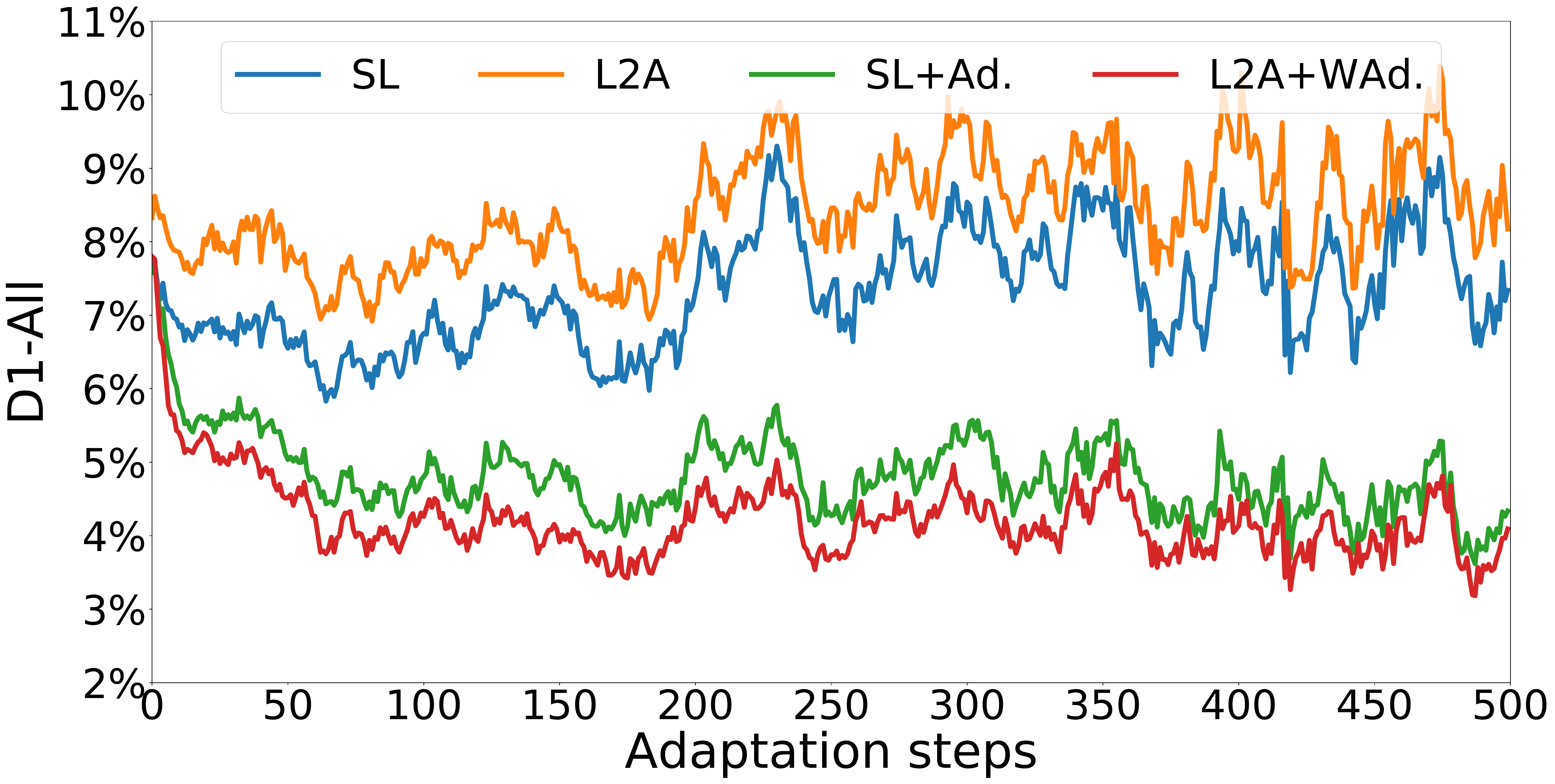}
	\caption{Average D1-all error with respect to the number of adaptation steps performed on the \kitti{} database by a \dispnet{} network trained according to supervised learning (\sul) or our learning to adapt framework (\lt).}
	\label{fig:meta_plot_time}
\end{figure}

\paragraph{Adaptation Performance Over Time:} 
To further highlight the difference of behaviour between models trained to regress and those trained to adapt, we plot the average {$\text{D1-all}$} error achieved by \dispnet{} on \kitti{} as a function of the number of adaptation steps in \autoref{fig:meta_plot_time}.
%The horizontal axis corresponds to the adaptation steps performed and to the frame number. 
The vertical axis represents the average D1-all error of the $k^\text{th}$ frame in all of the sequences in \kitti{}. 
Comparing the methods with and without online adaptation, it is clear that in both cases, adaptation drastically improves the performance. 
The comparison between \sula{} (green line) and \wlta{} (red line) shows how quickly our method adapts to the given video sequence. 
The poor results of \textbf{L2A} can easily be explained since our formulation never explicitly optimizes the base model for regression. Instead it optimizes the network to quickly learn-to-adapt, therefore the base model results can be sub-optimal, providing the performance can be improved in a few adaptation steps. 

\subsubsection{Synthetic to Synthetic}
\label{sssec:ablation}

\begin{table}[]
    \centering
    \scalebox{0.75}{
    \begin{tabular}{clccccc}
    
        \toprule%\cline{3-6}
        %\multicolumn{3}{c}{}&\multicolumn{2}{c}{Ad. Unsupervised}&\multicolumn{2}{c}{Ad. Supervised} \\
        %\midrule
        %&Method & Training Sets & D1-all (\%) & EPE & D1-all (\%) & EPE  \\
        &\multirow{2}{*}{Method}&\multirow{2}{*}{Training Set}&\multicolumn{2}{c}{Ad. Unsupervised}&\multicolumn{2}{c}{Ad. Supervised} \\
        & &  & D1-all (\%) & EPE & D1-all (\%) & EPE  \\
        \midrule
        (a) & \sula\cite{Tonioni2018real} & - & 26.56 & 3.96 & 15.60 & 2.24 \\
        \midrule
        (b) & \sula\cite{Tonioni2018real} & \scarla{} & 25.07 & 3.62 & 13.89 & 1.97 \\
        (c) & \lta{} & \scarla{} & 22.69 & 3.08 & \textbf{12.01} & \textbf{1.80} \\
        (d) & \wlta &  \scarla{} & \textbf{21.07} & \textbf{2.90} & \xmark & \xmark \\
        \bottomrule
    \end{tabular}
    }
    \vspace{3mm}
    \caption{Comparison of the training methods when evaluated on sequences from Synthia. It can be seen that the best performing training method is \wlta. We also provide results for when we use a $L_1$ supervised adaptation loss. Best results in bold.}
    \label{tab:carla_ablation}
\end{table}

Here, we perform a more controlled synthetic-to-synthetic evaluation where we can measure the difference in performance more explicitly thanks to the availability of dense and accurate ground truth labels. % \ie, train the models on one synthetic dataset and deploy them on another using online adaptation to reduce loss in performance. 
%The use of a synthetic dataset at test time has the key advantage of allowing us to benchmark more precisely performance differences between methods thanks to the availability of dense and accurate ground-truth data, rather than the sparse one of \kitti{}. 
The aim of the following series of tests will be to quantify the performance of the two key aspects of the learning-to-adapt framework, namely, learning to adapt through meta-learning and learning to weight noisy loss estimation. To further prove the generality of our learning to adapt formulation, we also provide results when the networks are trained to perform online adaptation using a supervised $L_1$ loss (\ie, ${\calL_u \equiv\calL_s}$).
%The aim of the following set of tests will be to quantify the contribution 

For these tests, we again use \dispnet{} trained on \scarla{} but tested on all the sequences of the full \ssynthia{} dataset.
%To compare different training strategies we have used the \dispnet{} network, pre-train it on \flying{}, fine-tuned it on \scarla{} according to different training strategies and tested it on the full \synthia{} dataset. 
%For all tests, we perform either unsupervised or supervised online adaptation. 
%In the first case, adaptation is performed minimizing the reprojection based loss described before, while for supervised adaptation we minimize the $L_1$ loss between the predicted disparity and the ground truth annotations online for each incoming stereo pair.
Specifically, to show that we can adapt using different loss functions, we train for both unsupervised and supervised adaptation\footnote{In online adaptation we use the $L_1$ loss between the predicted disparity and the ground truth annotations for each stereo pair.}, and evaluate the performance of the following training scenarios: (a) Using the initial model trained using \flying{}; (b) Training on \carla{} using a supervised $L_1$ loss; (c) Using the learning-to-adapt framework \textbf{without} confidence weighted loss; (d) Using the learning-to-adapt framework \textbf{with} confidence weighted loss.

We report the results in \autoref{tab:carla_ablation}, where it can be seen that explicitly training \dispnet{} to adapt using our \ltl{} formulation (c), allows the network to exploit the online adaptation and greatly improve the performance both in the unsupervised and supervised adaptation setups.
Finally, it can also be seen that weighting the unsupervised loss values results in superior performance (d). 
For this test set up, the results clearly demonstrate how our formulation is able to learn a weight configuration that is more inclined to be adapted to a new environment.

\subsection{Confidence Weighted Loss Function}
\label{ssec:confidence_qualitative}

\begin{figure}
	\centering
	\scalebox{0.90}{
	\begin{tabular}{c|c}
		\scarla{} & \kitti{} \\
		\hline
		\multicolumn{2}{c}{(a) Left RGB Frame}\\
		\includegraphics[width=0.25\textwidth]{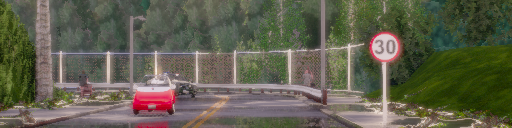} & \includegraphics[width=0.25\textwidth]{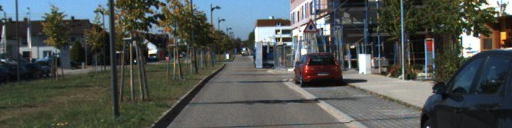} \\
		\multicolumn{2}{c}{(b) Disparity Predicted}\\
		\includegraphics[width=0.25\textwidth]{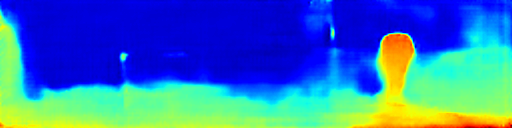} & \includegraphics[width=0.25\textwidth]{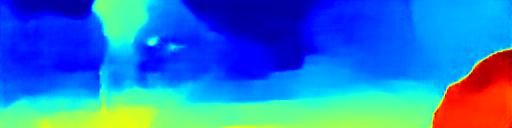} \\
		\multicolumn{2}{c}{(c) Reprojection Error ($\varepsilon$)}\\
		\includegraphics[width=0.25\textwidth]{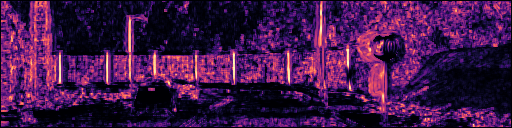} & \includegraphics[width=0.25\textwidth]{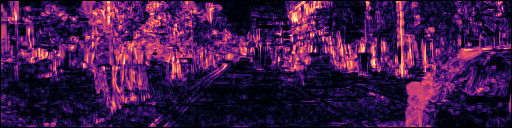} \\ 
		\multicolumn{2}{c}{(d) Confidence Mask ($W$)}\\
		\includegraphics[width=0.25\textwidth]{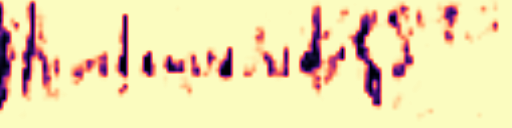} & \includegraphics[width=0.25\textwidth]{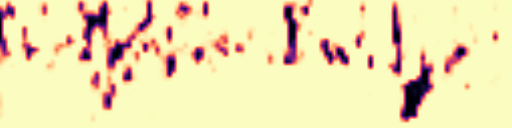} \\
		\multicolumn{2}{c}{(e) $W \odot \varepsilon$}\\
		\includegraphics[width=0.25\textwidth]{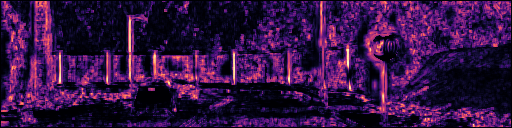} & \includegraphics[width=0.25\textwidth]{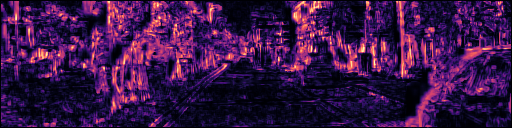} \\
	\end{tabular}
	}
	\vspace{0.5ex}
	\caption{Visualization of the errors optimized to achieve unsupervised adaptation with reprojection based loss function and using our weighting function. Brighter colours encode higher values.}
	\label{fig:meta_confidence}
\end{figure}

In \autoref{fig:meta_confidence}, we show a visualization of the confidence masks and weighted errors optimized by our confidence guided adaptation loss described in \autoref{ssec:method_reweighting}.
A quantitative evaluation is not possible due to the unavailability of the corresponding ground-truth data and obtaining it is not straightforward.
%as the ground-truth data  is non-existent and obtaining one  a non-trivial process.
% (before even considering the additional difficulties in objectively annotating problematic areas such as occlusions, texture-less regions or reflective surfaces). %(in addition to the difficulties introduced by attempting to objectively annotate problematic areas such as occlusions, texture-less regions or reflective surfaces).
%
%On the left column, we report samples obtained on the \carla{} dataset by a network trained on the same dataset, on the right samples from \kitti{} predicted by the same network.
%Row (c) shows the errors that will be optimized with the standard reprojection loss used for online adaptation in \cite{Tonioni2018real}. 
%Row (d) depicts the masks produced without any kind of direct supervision by our weighting network. 
The predicted confidence maps effectively mask out occluded regions in the image while keeping the useful error signals in the rest of the image (low confidence areas are encoded as dark pixels). 
%Row (e) shows the cleaner error estimation that will be optimized in our framework after weighting the loss estimated according to the predicted confidence mask. 
Errors on occluded regions, \eg, to the left of the traffic sign in the left column or to the left of the car in the right column, are effectively masked out, producing a cleaner error estimation that will improve adaptation performances. 
We wish to highlight that the confidence network has been trained without any direct supervision and only on \scarla{}, nevertheless, it seems to be able to generalize well to \kitti{}. 
We believe this ability to generalize is mainly due to the avoidance of working directly with RGB inputs, which inevitably change drastically between datasets.
Instead, the confidence network relies on the estimated re-projection error, which is more consistent across different environments.  

%%%%%%%%%%%%%%%%%%%%%%%%%%%%%%%%%%%%%%%%%%%%%%%%%%%%%%
\section{Discussion}
\label{sec:conclusions}
We have introduced a learning to adapt framework for stereo and demonstrated how the performance of deep stereo networks can be improved by explicitly training the network to be suited for adaptation. 
Moreover, we are able to automatically learn an implicit confidence measure, for noisy unsupervised error estimation, directly in our learning-to-adapt framework. 
Specifically, we showed  the ability of a \dispnet{} \cite{mayer2016large} network to adapt to a real and synthetic domain, when training is performed on a different synthetic domain.
In this setting, we obtained increased performance when applying our learning-to-adapt formulation. 
In future, we plan to test this framework on more complex network architectures (\eg, \cite{Kendall_2017_ICCV, zhong2018open}) and to extend it to use different unsupervised loss functions for online adaptation (\eg, the improved re-projection loss described in \cite{zhang2018activestereonet}). 

\section{Acknowledgement}
This work was supported by the ERC grant ERC-2012-AdG 321162-HELIOS, EPSRC grant Seebibyte EP/M013774/1, EPSRC/MURI grant EP/N019474/1 and TOSHIBA Research. We would also like to acknowledge the Royal Academy of Engineering and FiveAI.

{\small
\bibliographystyle{ieee}
\bibliography{biblio}
}

\newpage\phantom{blabla}


\section*{Supplementary material}

# Supplementary material for Learning to Adapt for Stereo

Alessio Tonioni[*1], Oscar Rahnama[†2,4], Thomas Joy[†2], Luigi Di Stefano[1], Thalaiyasingam Ajanthan[*3], and Philip H. S. Torr[2]

[1]University of Bologna  [2]University of Oxford  [3]Australian National University  [4]FiveAI

In Sec. 1 we provide additional implementation details concerning the loss used for unsupervised online adaptation, the confidence network and the training procedure of the disparity estimation network. Later in Sec. 2 we give more experimental results on various components of our method that further confirm the findings of the main paper.

## 1. Implementation Details

### 1.1. Unsupervised Online Adaptation Loss

To perform unsupervised online adaptation we use a modified version of the re-projection loss described in [3]. Denoting with $(i^l, i^r)$ an RGB stereo image pair with resolution $W \times H$, which is also normalized such that each pixel lies in the range $[0, 1]$. We use this pair as an input for the disparity estimation network which produces a disparity map $y^l$ aligned with $i^l$ as an output.

To compute the unsupervised adaptation loss, first, we use the scaled right input image $(i^r)$ and the disparity map $(y^l)$ to generate a re-projected left input image $(\tilde{i}^l)$ through differentiable bilinear sampling [3]. The error $\varepsilon_{ij}$ computed for any given pixel at coordinates $(i, j)$ is measured as a weighted sum of SSIM [7] (measured over $3 \times 3$ patches centred on $(i, j)$) together with an element-wise $L_1$ distance between real and reprojected pixels. Thus, the pixel error $\varepsilon_{ij}$ can be formally expressed as:

$$\varepsilon_{ij} = \alpha \frac{1 - \text{SSIM}\left(i^l_{ij}, \tilde{i}^l_{ij}\right)}{2} + (1 - \alpha) \left\| i^l_{ij} - \tilde{i}^l_{ij} \right\|. \quad (6)$$

We denote with $\alpha = 0.85$ the weight of the linear combination of the two losses. The final loss optimized to perform online adaptation is the mean re-projection error across all the pixels of the whole image.

### 1.2. Confidence Network Architecture

We implement the confidence estimation network as a three layer 2D convolutional neural network without pooling. Each of these layers uses $3 \times 3$ kernels and a stride of 1. The first and second layers have 32 and 64 filters, respectively, and both make use of leaky ReLU activations as well as batch normalization. The final layer has a single channel output and uses the sigmoid function as activation.

We use the re-projection errors (estimated according to Eq. 6) as an input to our confidence network. Before being fed into the confidence network, this re-projected error map is first downsampled to a quarter of its original size to reduce memory consumption and increase inference speed. With this, the network then outputs a confidence map with the same reduced dimensions which we then upscale back to full resolution using bilinear interpolation.

### 1.3. Training Details

All our systems are entirely implemented in TensorFlow. The pre-training of Dispnet is performed for 1200K steps on *F3D* from random initialization using Adam optimizer with an initial learning rate of 0.0001. The learning rate following the authors' guidelines is halved at 600K steps and then again at 1000K steps. The network is trained by minimizing a weighted sum of the multiple losses computed at different output resolutions. We refer the interested reader to [4] for additional details on the scheduling of the weights associated with loss at different resolutions across the training process.

Fine tuning of both the baselines and our methods is performed for 40K steps on the training dataset with the hyperparameters that are detailed in the main paper. For these fine tunings the loss functions are computed only for prediction at full resolution.

## 2. Additional Results

### 2.1. Synthetic to Synthetic Experiments

We extend the experiment performed in Sec. 5.2.2 in the main paper by swapping the role of the two synthetic datasets and report the results in Tab. 3. Here we perform training of different methods on Synthia [5] and test the

---
[*]Work done while at University of Oxford.
[†]Second two authors contributed equally.

1

|     | Method    | Training Set | Ad. Unsupervised D1-all (%) | EPE  | Ad. Supervised D1-all (%) | EPE  |
| --- | --------- | ------------ | --------------------------- | ---- | ------------------------- | ---- |
| (a) | **SL+Ad** | -            | 16.63                       | 2.56 | 10.56                     | 1.50 |
| (b) | **SL+Ad** | Synthia      | 12.53                       | 1.71 | 5.36                      | 0.87 |
| (c) | **L2A+Ad**| Synthia      | 10.79                       | 1.48 | **5.09**                  | **0.85** |
| (d) | **L2A+WAd**| Synthia     | **9.98**                    | **1.45** | ✗                     | ✗    |

Table 3. Comparison of the different methods when training on Synthia and evaluating on sequences from Carla. Similar to our previous results in Tab. 2, the best performing training method is **L2A+WAd**. We also provide results when a $L_1$ based supervised adaptation loss is used at test time. Best results are in bold.

| Method    | Train on Carla D1-All (%) | EPE  | Train on Synthia D1-All (%) | EPE  |
| --------- | ------------------------- | ---- | --------------------------- | ---- |
| **L2A+Ad** | 22.69                    | 3.08 | 10.79                       | 1.48 |
| **FOL2A+Ad** | 23.14                 | 3.12 | 10.81                       | 1.49 |

Table 4. Comparison between our full framework and a first-order approximation of it. All the models are trained on one of the two synthetic datasets and tested on the other, by performing unsupervised online adaptation.

models performing online adaptation on sequences from Carla. We report results for both unsupervised and supervised online adaptation for the base model trained on *F3D* (row (a)), the model fine-tuned according to an $L_1$ regression loss (row (b)), and our learning to adapt framework with and without the confidence network (row (d) and (c)) respectively.

As in the main paper, our **L2A+Ad** formulation provides an increase in performance over **SL+Ad** for both unsupervised and supervised adaptation. Moreover, the introduction of our confidence weighted adaptation (row (d)) provides an additional small increase in performance for unsupervised adaptation. In this scenario, however, the gap between the different methods appears to be smaller.

## 2.2. First-Order Approximation

In MAML [2], the use of two nested optimization loops introduces the need for second-order derivatives during the back-propagation phase through gradient computations of the inner loop. Naturally, the computation of second-order derivative comes at a significant computational cost that slows down the training process considerably. To alleviate this issue, the authors of MAML [2] propose a first-order approximation that ignores the costly back-propagation through the gradient computation steps and leads to similar performance at a more affordable computational cost.

Our learning to adapt formulation, described in Sec. 3.1 of the main paper, also uses two nested optimizations, and therefore may benefit from the same kind of approximation. This approximated version can be easily implemented in our framework exactly as in MAML by omitting the computation of the costly second order derivatives during the back propagation. This approximated method is referred to as **FOL2A**.

To measure the impact of the first order approximation on the final performance of the network we trained two additional Dispnet using **FOL2A** on a synthetic dataset (Carla or Synthia) and measured the performance on the other (*i.e.*, we followed the paradigm used for the synthetic to synthetic tests) . In Tab. 4 we compare the performance achieved by the networks trained with **FOL2A** and the corresponding models trained without approximations (*i.e.*, using **L2A**). On both training datasets, the approximated methods perform similarly to the complete ones, with performance almost equal when trained on Synthia. Hence, if one wishes to save on computational resources, this first order approximation approach may be employed.

Unfortunately, weighted adaptation (using the confidence network) is not possible with the first-order approximation as the confidence network is trained using the gradients computed through the adaptation step (inner loop) and these gradients are ignored in the first-order approximation. Therefore, for the sake of fairness, when comparing between the different variants of our method across all the tests, we have always used the version that relies on the computation of the second-order derivatives.

## 2.3. Qualitative Results on the Confidence Network

In Fig. 6, we show additional qualitative examples of our predicted confidence masks on a challenging sequence from the KITTI dataset. We wish to point out that, in our learning-to-adapt framework, despite the confidence network only being trained on synthetic data, it demonstrates good generalization ability to real images. In almost all examples, the confidence network is able to mask out reprojection errors due to occlusions (usually on the left side of objects in the foreground) or reflective surfaces (*e.g.*, the car on the left in the second row). In some examples, we note that certain points in the background are masked. This slightly unexpected behavior may hint to the presence of noise within the confidence prediction process. Future work

will be dedicated to improve the quality of these confidence estimations by incorporating geometric constraints.

### 2.4. Qualitative Results on Cityscapes

We show on Fig. 7 some additional qualitative results obtained on the Cityscapes[1] dataset. We show three different disparity prediction obtained using Dispnet as disparity estimation network and three different adaptation strategies; from left to right: no adaptation (**SL**), online adaptation as in [6] (**SL+Ad**) and our proposed framework (**L2A+WAd**). The number of adaptation steps performed increase by roughly 25 together with the row considered. The visualization clearly shows how both adaptation strategies (+Ad. methods) are able to address most of the nuisances in the prediction of **SL**, *i.e.* those produced by a method without adaptation. The differences between **SL+Ad** and **L2A+WAd** are quite evident in the first row, *i.e.* after few step of adaptation, but became quite subtle by the last row. Nevertheless, **L2A+WAd** is always able to obtain sharper and cleaner prediction.

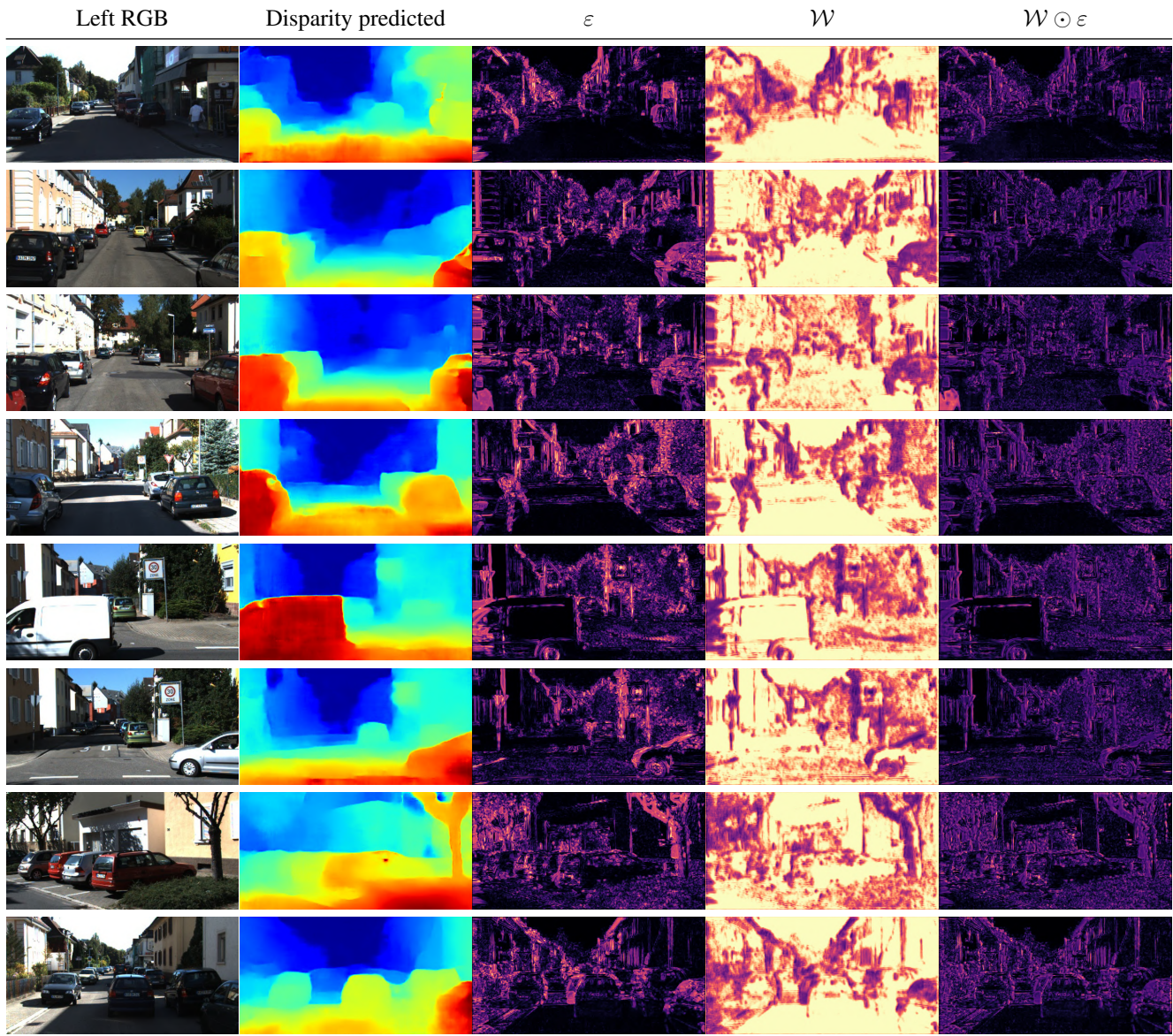

Figure 6. Visualization of our confidence estimations. From left to right: left rgb frame from a stereo pair, disparity predicted by Dispnet, reprojection error obtained as described in Sec. 1.1, confidence mask predicted by our network, weighted re-projection error used for online adaptation. On the last three columns bright colors indicate high values.

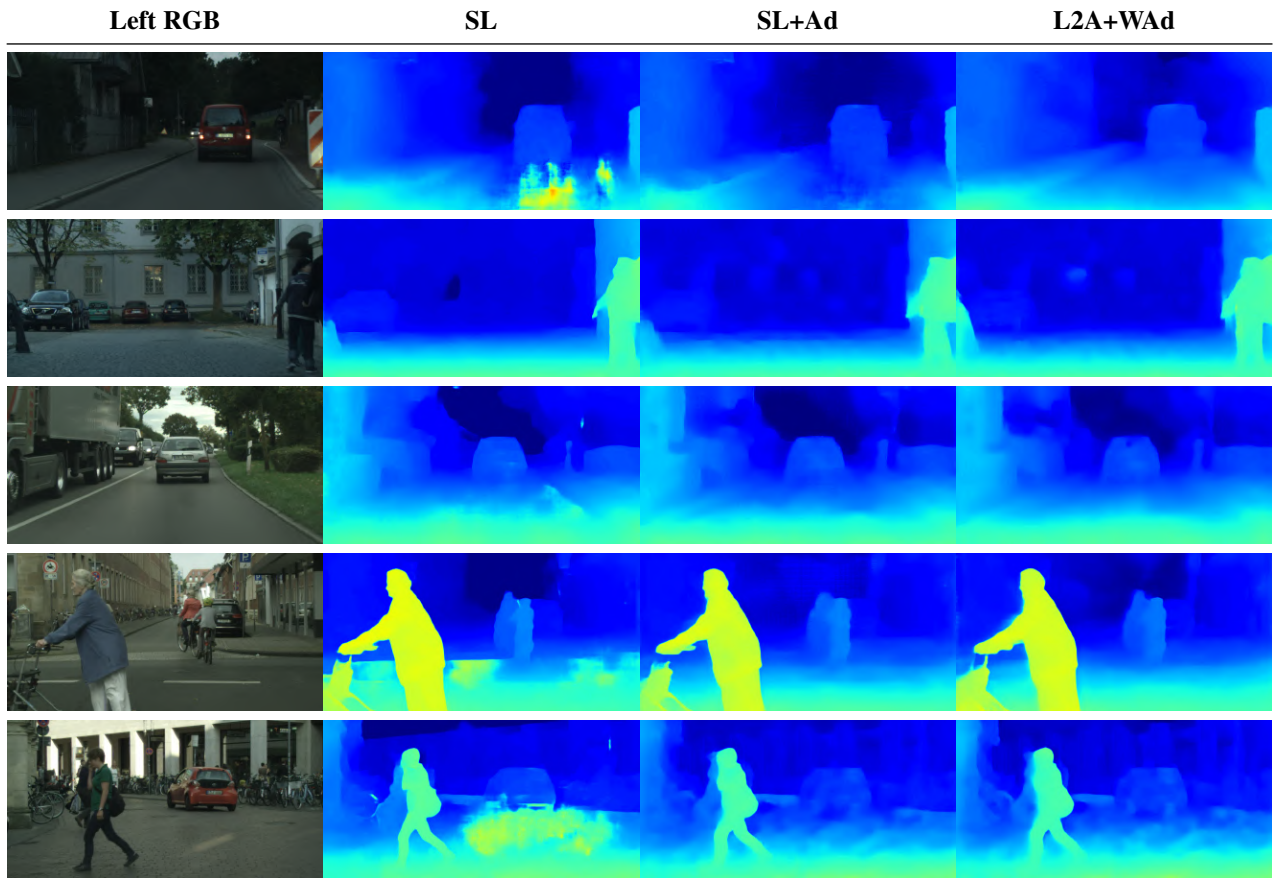

Figure 7. Qualitative results on the Cityscapes dataset. From left to right: left rgb frame from a stereo pair, disparities predicted by Dispnet using as method **SL**, **SL+Ad** and **L2A+WAd** respectivelly. On the last three columns bright colors marks high disparity values. The number of adaptation steps performed increase together with the rows.



\end{document}